\documentclass[10pt,twocolumn,letterpaper]{article}

\usepackage{cvpr}              

\usepackage{graphicx}
\usepackage{amsmath}
\usepackage{amssymb}
\usepackage{booktabs}
\usepackage{multirow}
\makeatletter
\@namedef{ver@everyshi.sty}{}
\makeatother
\usepackage{tikz}

\usepackage[pagebackref,breaklinks,colorlinks]{hyperref}

\usepackage[capitalize]{cleveref}

\newcommand{\tabincell}[2]{\begin{tabular}{@{}#1@{}}#2\end{tabular}}
\newcommand*\circled[1]{\tikz[baseline=(char.base)]{\node[shape=circle,draw,inner sep=0.8pt] (char) {#1};}}

\crefname{section}{Sec.}{Secs.}
\Crefname{section}{Section}{Sections}
\Crefname{table}{Table}{Tables}
\crefname{table}{Tab.}{Tabs.}

\def\cvprPaperID{4} 
\def\confName{CVPR}
\def\confYear{2022}

\begin{document}

\title{Democratizing Contrastive Language-Image Pre-training: \\A CLIP Benchmark of Data, Model, and Supervision}

\author{Yufeng Cui$^{1, 3 \footnotemark[1]}$, Lichen Zhao$^{1,3 \footnotemark[1]}$, Feng Liang$^{2 \footnotemark[1]}$, Yangguang Li$^{3 \footnotemark[2]}$, Jing Shao$^3$\\
$^1$ Beihang University, $^2$ University of Texas at Austin, $^3$ SenseTime Research\\
{\tt\small \{cuiyufeng, zlc1114\}@buaa.edu.cn, jeffliang@utexas.edu }\\
{\tt\small \{liyangguang,shaojing\}@sensetime.com }
}

\maketitle

\renewcommand{\thefootnote}{\fnsymbol{footnote}} 
\footnotetext[1]{Equal contribution.} 
\footnotetext[2]{Corresponding Author.} 

\begin{abstract}
Contrastive Language-Image Pretraining (CLIP) has emerged as a novel paradigm to learn visual models from language supervision. 
While researchers continue to push the frontier of CLIP, reproducing these works remains challenging. 
This is because researchers do not choose consistent training recipes and even use different data, hampering the fair comparison between different methods.
In this work, we propose CLIP-benchmark, a first attempt to evaluate, analyze, and benchmark CLIP and its variants. 
We conduct a comprehensive analysis of three key factors: data, supervision, and model architecture.
We find considerable intuitive or counter-intuitive insights: 
(1). Data quality has a significant impact on performance. 
(2). Certain supervision has different effects for Convolutional Networks (ConvNets) and Vision Transformers (ViT). 
Applying more proper supervision can effectively improve the performance of CLIP.
(3). Curtailing the text encoder reduces the training cost but not much affect the final performance.
Moreover, we further combine DeCLIP~\cite{li2021supervision} with FILIP~\cite{yao2021filip}, bringing us the strongest variant DeFILIP. The CLIP-benchmark would be released at: \url{https://github.com/Sense-GVT/DeCLIP} for future CLIP research.

\end{abstract}

\section{Introduction}

Over the past few years, supervised pre-training on well-labeled ImageNet~\cite{deng2009imagenet} and then transferred to downstream tasks~\cite{girshick2014rich, long2015fully, vinyals2015show} has greatly transformed the computer vision (CV) community. 
However, supervised pre-training is hard to scale since we need arduous human labeling to specify new visual concepts.
More Recently, Contrastive Language-Image Pretraining (CLIP)~\cite{radford2021learning} has emerged as a scalable pre-training paradigm via learning visual models from language supervision, or more specifically, image-text pairs.
Basically, CLIP adopts the contrastive loss to push the embeddings of matched image-text pairs together while pushing those of non-matched pairs apart.
Benefiting from abundant image-text pairs on the Internet, CLIP learns general visual features that could perform zero-shot recognition, \ie, predict an image's category without seeing a single labeled example.
CLIP's transferable features could also be well transferred to various downstream tasks.

\begin{table}[t]
\begin{center}
\caption{\label{tab: clip_variants} 
Summary of CLIP and its variants. 
Although several approaches use the same amount of 15 million data from YFCC, the filtering strategies are different. 
V1$^\dagger$ is filtered by CLIP. V2$^\star$ is filtered by DeCLIP.}
\small
\begin{tabular}{cccc}
  \toprule
  \multirow{2}{*}{\textbf{Method}} & \textbf{Training} & \textbf{Available} & \textbf{Image}  \\
   & \textbf{code} & \textbf {data} & \textbf{encoder} \\
  \midrule   
  CLIP~\cite{radford2021learning} 	& No & YFCC15M-V1$^\dagger$ & ViT, ResNet\\
  SLIP~\cite{mu2021slip} & Yes & YFCC15M-V1$^\dagger$ & ViT \\
  DeCLIP~\cite{li2021supervision} & Yes & YFCC15M-V2$^\star$ & ViT, ResNet\\
  FILIP~\cite{yao2021filip} & No & -- & ViT\\
  \bottomrule 
\end{tabular}
\end{center}
\end{table}

Witnessing its great success, researchers continue to push the frontier of CLIP. 
For instance, SLIP~\cite{mu2021slip}, DeCLIP~\cite{li2021supervision} and FILIP~\cite{yao2021filip} achieve considerable improvements via embracing different kinds of supervision within the image-text pairs.
However, it remains challenging to make fair comparison between these methods.
This is because they do not choose consistent training recipes and even use different data.
As we can see from Tab.~\ref{tab: clip_variants}, although CLIP~\cite{radford2021learning}, DeCLIP~\cite{li2021supervision} and SLIP~\cite{mu2021slip} use the same amount of 15 million data from YFCC~\cite{thomee2016yfcc100m}, they adopt different filtering strategies.
Moreover, methods~\cite{radford2021learning,jia2021scaling,li2021supervision,yao2021filip,pham2021combined} crawl their datasets from the Internet, making the fair comparison more difficult.

This paper aims to democratize large-scale CLIP, \ie, to build a fair and reproducible CLIP community.  
We propose CLIP-benchmark, a first attempt to evaluate, analyze, and benchmark CLIP and its variants. 
We do a comprehensive empirical study on three key factors: data, supervision, and model architecture. 
We find considerable intuitive or counter-intuitive insights:

\begin{itemize}

    \item \textbf{Data}: Mid-scale 15M data is a good balance of the training cost and performance. 
    Thus, most methods use a 15M subset from YFCC~\cite{thomee2016yfcc100m} to verify the effectiveness of their methods. 
    We carefully compare the current two YFCC15M versions, V1 from CLIP~\cite{radford2021learning} and V2 from DeCLIP~\cite{li2021supervision}. 
    Interestingly, we find that in terms of the zero-shot performance on ImageNet, V2 is much better than V1 (details in Tab.~\ref{tab: quality}). 
    We conjecture that V2 includes a more meticulous filtering strategy, making its data quality better than V1.
    This also helps us conclude that data quality is crucial in CLIP training.
    
    \item \textbf{Supervision}: We first reproduce all the methods using a unified training recipe (details in Tab.~\ref{tab:ablation on YFCC}). 
    We find that fine-grained alignment supervision~\cite{yao2021filip} could benefit ViT image encoder but hurts ConvNets.
    Intuitively, fine-grained alignment needs the image features to be non-overlapped, which is unachievable for ConvNets.
    For ViT image encoder, aggregating self-supervision~\cite{mu2021slip, li2021supervision}, multi-view supervision~\cite{li2021supervision}, nearest-neighbor supervision ~\cite{li2021supervision} and fine-grained alignment supervision~\cite{yao2021filip} brings us the strongest variant DeFILIP.
    
    \item \textbf{Model}: While most attention is paid to image encoders, little research is conducted on text encoders. 
    Most literature follows the exact setting from CLIP, \ie, a 12-layer transformer~\cite{radford2019language}.
    We find that CLIP's text encoder is not necessary to be so much deep; a 3-layer transformer performs even better than the default 12-layer setting under the mid-scale data scenarios (details in Tab.~\ref{text-encoder-size}).
    Therefore, pay attention to the text encoder when designing your CLIP models.
    
\end{itemize}

In a nutshell, this paper proposes the first CLIP-benchmark that includes the state-of-the-art methods. We benchmark these methods under the same training recipe using the same data. Our CLIP-benchmark also brings some insights about data, supervision and model. The CLIP-benchmark would be released to the public for future research.

\section{Related Work}

Concurrently to this work, many researchers continue to push the frontier of CLIP~\cite{radford2021learning}. SLIP~\cite{mu2021slip} introduces self-supervision to Contrastive Language-Image Pretraining. 
DeCLIP~\cite{li2021supervision} utilizes widespread supervision among the image-text pairs. 
FILIP~\cite{yao2021filip} leverages the finer-grained alignment between image patches and textual words.
LiT ~\cite{zhai2021lit} adopt contrastive-tuning to tune the text tower using image-text data while using a pre-trained, strong image model as the image tower. 
OTTER~\cite{wu2021data} uses online entropic optimal transport to find a soft image-text match as labels for contrastive learning. 

The representations learned by CLIP have shown excellent transferability over various tasks.
CLIP2Video~\cite{fang2021clip2video} and CLIP4Clip~\cite{luo2021clip4clip} apply CLIP to video retrieval task. 
ActionCLIP~\cite{wang2021actionclip} utilizes CLIP for action recognition task. 
More works about improving image captioning with CLIP, \eg CLIPCap~\cite{mokady2021clipcap}, CLIP4Caption~\cite{tang2021clip4caption}. 
Interestingly CLIP can be even used in text-guided image generation task ( StyleCLIP~\cite{patashnik2021styleclip}) and Embodied AI ( EmbCLIP~\cite{khandelwal2021simple}). CLIP has also contributed to the development of general vision~\cite{shao2021intern}. Witnessing CLIP's active community and wide applications, we propose the first work to benchmark CLIP.

\section{Methods}

\begin{figure*}[ht]
    \centering
	\includegraphics[width=1.8\columnwidth]{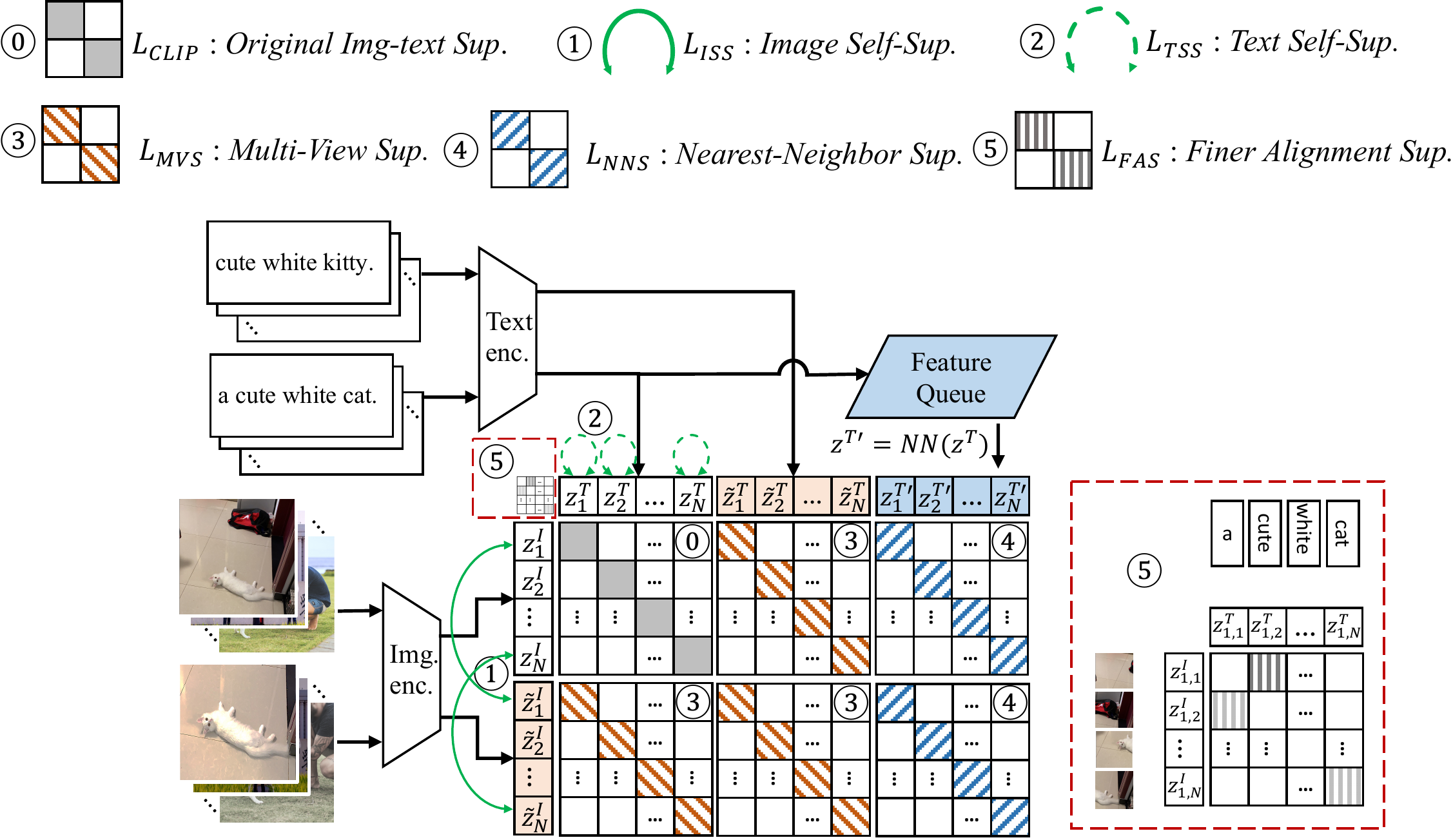}
	\caption{\label{fig:clip_framework} A unified framework of CLIP variants. 
	Combining different supervision leads to different variants. CLIP: \{\protect\circled{0}\}, SLIP: \{\protect\circled{0},\protect\circled{1}\}, FILIP: \{\protect\circled{5}\}, DeCLIP: \{\protect\circled{0},\protect\circled{1},\protect\circled{2},\protect\circled{3},\protect\circled{4}\}, DeFILIP: \{\protect\circled{0},\protect\circled{1},\protect\circled{2},\protect\circled{3},\protect\circled{4},\protect\circled{5}\},}
\end{figure*}

CLIP~\cite{radford2021learning} and its variants (\eg, DeCLIP~\cite{li2021supervision}, SLIP~\cite{mu2021slip}, and FILIP~\cite{yao2021filip}) follow a common high-level structure(see Fig.~\ref{fig:clip_framework}). 
The model consists of an image encoder (\eg, ResNet~\cite{he2016deep} or ViT~\cite{dosovitskiy2020image}) and a text encoder(\eg, transformer~\cite{vaswani2017attention}), with a multimodal interaction at the top. 
Take the most straightforward CLIP as an example, the image encoder (the text encoder) extracts the image embedding (the text embedding) based on the input image-text pair. 
A contrastive objective is used to push the embeddings of matched image-text pairs together while pushing non-matched pairs apart. 
At the test phase, the learned text encoder synthesizes a zero-shot linear classifier by embedding the arbitrary categories of the test dataset. 
As shown in the Fig.\ref{fig:clip_framework}, different variants further explore the widespread supervised signal of the image-text pair for better visual representations. 
This section will briefly introduce the above CLIP variants and bring the strongest variant DeFILIP.

\subsection{CLIP}

CLIP~\cite{radford2021learning} only uses the original image-text supervsion. In a batch of $N$ image-text pairs ${\{(x_i^I, x_i^T)\}}$, we denote $ x_i^I$ and $x_i^T$ as image and text of the $i_{th}$ pair. Let $z_i^I$ and $z_j^T$ be the normalized embedding of the $i_{th}$ image and $j_{th}$ text, respectively. CLIP uses InfoNCE loss~\cite{van2018representation}. The loss for the image encoder can be denoted as Eq.~\ref{eq:infonce}.

\begin{equation}
\label{eq:infonce}
    L_I = - \frac{1}{N} \sum_{i=1}^{N} \log \frac{\exp(\mathrm{sim}(z_i^I,  z_i^T)/\tau)}{\sum_{j=1}^{N}\exp(\mathrm{sim}( z_i^I,  z_j^T)/\tau)}~
\end{equation}

Here, the similarity function $\mathrm{sim}(,)$ is measured by dot product, and $\tau$ is a learnable temperature variable to scale the logits. We have a symmetrical loss for image and text encoder; thus, the overall loss function $L_{CLIP}$ is the average of $L_I$ and $L_T$.

\begin{equation}
L_{CLIP}= (L_I + L_T)/2
\end{equation}

At the test phase, the learned text encoder synthesizes a zero-shot linear classifier by embedding the arbitrary categories of the test dataset. Because it is rare in the dataset that image caption is just a single word, CLIP uses prompts to make up the context of the category \texttt{\{label\}}, such as \texttt{"a photo of a \{label\}"}. 

\subsection{SLIP}

SLIP ~\cite{mu2021slip} introduces self-supervision to CLIP for better visual representations. 
Built upon CLIP, SLIP gets two more strong augmented views for image self-supervised contrastive loss $L_{ISS}$. SLIP further compares different image self-supervised methods and finally selected SimCLR~\cite{chen2020simple} for the final framework. The overall loss function of SLIP is shown in Eq.~\ref{eq:slip_loss}. $\alpha$ is the scale of self-supervision and is set to 1.

\begin{equation}
\label{eq:slip_loss}
    L_{SLIP} =L_{CLIP} + \alpha L_{ISS}
\end{equation}


\subsection{FILIP}

FILIP ~\cite{yao2021filip} perform finer-grained alignment supervision on token level rather than image-text level. The similarity ${\{sim( z_i^I, _i^T)\}}$ of the $i_{th}$ image and $j_{th}$ text is improved to token-wise maximum similarity which is calculated as:

\begin{equation}
\label{eq:filip_loss}
\begin{cases}
    \mathrm{sim^I}(z_i^I, z_j^T) = \frac{1}{n_1} \sum_{k=1}^{n_1} z_{i,k}^I z_{j,m_k^I}^T \\
  
    \mathrm{sim^T}( z_i^I, z_j^T) = \frac{1}{n_2} \sum_{k=1}^{n_2} z_{i,m_k^T}^I z_{j,k}^T \\
\end{cases}
\end{equation}

Where $ m_k^I = argmax_{0<r<n_2} z_{i,k}^I z_{j,r}^T $ and $m_k^T = argmax_{0<r<n_1} z_{i,r}^I z_{j,k}^T $. FILIP achieves finer-level alignment through a cross-modal late interaction mechanism, which uses a token-wise maximum similarity between visual and textual tokens to guide the contrastive objective. 
Though the late cross-modal interaction can capture finer-grained features, it relies on the token-wise representations of both modalities and can be inefficient in terms of communication, memory, and computation. To alleviate this problem, the authors carefully reduce the precision and embedding size of the model and further select the 25\% tokens with the highest token-wise maximum similarity score among all texts (\emph{resp}, images) in the same local worker before node communication. Denoting the loss of fine-grained alignment supervision as $L_{FAS}$, The overall loss function of FILIP is shown in Eq.~\ref{eq:filip_loss}.

\begin{equation}
\label{eq:filip_loss}
    L_{FILIP} = L_{FAS}
\end{equation}


\subsection{DeCLIP}

DeCLIP ~\cite{li2021supervision} utilizes widespread supervision among the image-text pairs, including Self-Supervision(SS), Multi-View Supervision(MVS), and Nearest-Neighbor Supervision(NNS). DeCLIP contains image SS and text SS: Image SS maximizes the similarity between two augmented views of the same instance while text SS leverages Masked Language Modeling(MLM) within a text sentence.
For MVS, DeCLIP has two augmented views of both image and text, then contrasts the $2\times2$ image-text pairs. For NNS, DeCLIP sample text NN in the embedding space as additional supervision.

In summary, DeCLIP denote $L_{ISS}$ and $L_{TSS}$ as the loss function of image SS and text SS, respectively. $L_{MVS}$ is multi-view loss, and $L_{NNS}$ is nearest-neighbor loss. The overall loss function of DeCLIP is shown in Eq.~\ref{eq:declip_loss}. $\alpha, \beta, \gamma$ are the loss scales and are both set to 0.2. 

\begin{equation}
\label{eq:declip_loss}
\begin{aligned}
    L_{DeCLIP} = & (1 - \alpha - \beta - \gamma) L_{CLIP}  \\
    & + \alpha (L_{ISS} + L_{TSS}) \\
    & + \beta L_{MVS} +  \gamma L_{NNS} \\
\end{aligned}
\end{equation}

\subsection{DeFILIP}

By introducing the above methods, we can find a large number of possible supervision signals in the image-text pairs, which can improve the efficiency of training and generalization ability. In order to learn better visual representations and improve the data efficiency of the model, we further combine DeCLIP~\cite{li2021supervision} with FILIP~\cite{yao2021filip}, bringing us the strongest variant DeFILIP. 
The overall loss function of DeFILIP is shown in Eq.~\ref{eq:defilip_loss}.

\begin{equation}
\label{eq:defilip_loss}
\begin{aligned}
    L_{DeFILIP} = & (1 - \alpha - \beta - \gamma) L_{CLIP}  \\
    & + \alpha (L_{ISS} + L_{TSS}) \\
    & + \beta L_{MVS} +  \gamma L_{NNS} \\
    & + \lambda L_{FAS}
\end{aligned}
\end{equation}

$L_{FAS}$ is applied to improve fine-grained learning of visual representations further. The loss weight $\lambda$ is set to 0.2 in this work. As shown in fig. \ref{fig:clip_framework}, our DeFILIP is a summary and development of the existing SOTA methods, which   applies the existing supervision and achieves a new state-of-the-art performance.

\section{CLIP-Benchmark}

\subsection{Setup}

\paragraph{Evaluation Metric} In this paper, we mainly evaluate zero-shot performance of different models, which is regarded as the main feature of CLIP methods. 
We choose the zero-shot top-1 accuracy on ImageNet~\cite{deng2009imagenet} as the evaluation metric. 
We perform prompt ensemble by averaging the caption embeddings for each class across the prompt templates. 
The prompts are the same as proposed in CLIP\cite{radford2021learning}.

\paragraph{Implementation details}
The models in this work are trained and tested in the same codebase. Unless otherwise specified, all models are realized with Pytorch, and are trained with 32 NVIDIA A100 GPUs. When pretraining, we use an AdamW optimizer~\cite{loshchilov2017decoupled} with a total batch size of 4,096 (single GPU batch size 128). Starting with a learning rate (LR) of 0.0001, we linearly increase the LR to 0.001 (a.k.a warm-up) in one epoch, and then we use the cosine anneal LR decay strategy~\cite{loshchilov2016sgdr} to decrease the LR. The weight decay rate is set to 0.1.  The input resolution of the image encoder is 224 $\times$ 224, and the maximum context length of the text encoder is 76. The learnable temperature parameter $\tau$ in Eq.\ref{eq:infonce} is initialized to 0.07. We train the ResNet50 (\texttt{abbr}. as R50) and ViT-B/32 (\texttt{abbr}. as V-B32) from scratch for 32 epochs. 

\paragraph{Data augmentation}
SLIP, FILIP and DeCLIP all perform data augmentation for images during the pre-training phase. The augmentation policy includes: \texttt{RandomResizedCrop} with scale in [0.2,1.0]~\cite{wu2018unsupervised}, \texttt{ColorJitter} containing \{brightness, contrast, saturation, hue\} strength of \{0.4, 0.4, 0.4, 0.1\} with an applying probability of 0.8, \texttt{RandomGrayscale} with an applying probability of 0.2. Blurring augmentation~\cite{chen2020simple} has a Gaussian kernel with std in [0.1, 2.0], and \texttt{RandomHorizontalFlip}. Only DeCLIP performs data augmentation for texts. DeCLIP uses the EDA~\cite{wei2019eda} method as their text augmentation strategy, which contains three types of text augmentation strategies: synonym replacement, random swap, and random deletion.


\subsection{Data} 

Data is a crucial part of CLIP. This section does a holistic study of two mid-scale YFCC15M versions. V1 from CLIP~\cite{radford2021learning} and V2 from DeCLIP~\cite{li2021supervision}.

\vspace{-0.5em}
\paragraph{Data statistics}
We present statistics of two versions YFCC15M on examples number, mean/std of caption length, mean English word ratio, and the vocabulary size (unique tokens) in Table \ref{tab:yfcc15}. 
The V2 consists of 15.4M image-text pairs, 0.6M(3\%) more than V1.
V2 is generally shorter and more evenly distributed than V1 regarding the caption length. 
The English word ratio (\ie, \# of English words divided by \# of all words) of V2 is about 0.92, which is significantly better than V1's 0.72. 
For the vocabulary size (unique tokens), V1 is one order larger than V2 mainly because V1 contains many non-English characters.
We can infer from these statistics that V2 has better quality than V1 because V2 is more evenly distributed and has fewer non-English characters. 
We believe that V2 includes a more meticulous filtering strategy, making its data quality better than V1.

\setlength\tabcolsep{4pt}
\begin{table}[h]
\begin{center}
\begin{small}
\caption{\label{tab:yfcc15} The basic statistics of the two versions of YFCC15M. V1 is filtered by CLIP. V2 is filtered by DeCLIP.}
\vspace{-1.0em}
\begin{tabular}{ccccc}
  \toprule  
  \textbf{Dataset}  
  & \textbf{Examples}  
   & \tabincell{c}{\textbf{Caption} \\ \textbf{length}}
  & \tabincell{c}{\textbf{En-word} \\ \textbf{ratio}} 
  & \tabincell{c}{\textbf{Unique} \\ \textbf{Tokens}} \\
  \midrule   
  V2 & 15,388,848 & 16.7$\pm$29.2 & 0.92 & 770,996	\\
  V1 & 14,747,529 & 26.1$\pm$69.6 & 0.72 & 8,262,556	\\
  \bottomrule 
\end{tabular}
\label{sample-table}

\end{small}
\end{center}
\end{table}
\setlength\tabcolsep{6pt}

\noindent\textbf{Performance over V1-V2}. To further evaluate the quality of the two YFCC15M versions and explore the impact of data quality on CLIP, we perform a comparison with V1~\cite{radford2021learning} and V2~\cite{li2021supervision}, using the same methodology. As shown in Tab.~\ref{quality-table}, training with V2 leads to a better zero-shot performance than V1 under the same experimental setup.
On the one hand, it proves that the data quality of V2 is better regarding final performance. On the other hand, it also proves that data quality significantly impacts the performance of CLIP methods. 

\setlength\tabcolsep{3pt}
\begin{table}[t]
\begin{small}
\begin{center}
\caption{\label{tab: quality} Zero-shot top1 accuracy on ImageNet. We train CLIP-ViT-B32 and our DeFILIP-ViT-B32 using different datasets}
\vspace{-1.0em}
\begin{tabular}{cccc}
  \toprule
  \textbf{Method} & \textbf{Accuracy w/ V1} & \textbf{Accuracy w/ V2} \\
  \midrule   
  CLIP  & 26.1 & 32.8 \\
 DeFILIP & 36.4 & 45.0 \\
  \bottomrule 
\end{tabular}
\label{quality-table}
\end{center}
\vspace{-2.0em} 
\end{small}
\end{table}

\subsection{Supervision}


We perform a comprehensive comparison of our re-implemented pretraining methods \cite{radford2021learning, li2021supervision, mu2021slip, yao2021filip} to benchmark these methods under the same training recipe. 
We report the zero-shot top-1 accuracy on ImageNet in Tab.~\ref{tab:ablation on YFCC}
When the image encoder is ViT, all supervision is proved to be effective. DeCLIP, which utilizes the maximum supervision, obtains the best results. 
Moreover, we further integrate the existing supervision to make the strongest variant, named DeFILIP. Our proposed DeFILIP reaches 45.0\% accuracy, surpassing the CLIP baseline by a considerable 12.2\% margin.
 


 \begin{table}[h]
\caption{\label{tab:ablation on YFCC} Zero-shot top1 accuracy on ImageNet. All models are trained with YFCC15M-V2~\cite{li2021supervision}. 
$\Delta$ denotes the improvement. 
} 
\begin{center}
\vspace{-1.5em}
\begin{tabular}{cccccc}

\toprule 
\textbf{Method} & \textbf{Image encoder} & \textbf{Accuracy} & \textbf{$\Delta$} \\
\midrule
CLIP                             & \multirow{4}{*}{ResNet50}               & 37.2                               & -                                  \\
SLIP                             &                                         & 28.5                               & -                                  \\
FILIP                            &                                         & 21.3                               & -                                  \\
DeCLIP                           &                                         & \textbf{44.4}     & +7.2                               \\
\midrule
\hline
CLIP                             & \multirow{4}{*}{ViT-B/32}               & 32.8                               & -                                  \\
SLIP                             &                                         & 34.3                               & +1.5                               \\
FILIP                            &                                         & 39.5                               & +6.7                               \\
DeCLIP                           &                                         & \textbf{43.2}     & +10.4                              \\ \hline
DeFILIP                          & ViT-B/32                                & \textbf{45.0}     & +12.2                              \\ \bottomrule
\end{tabular}
\end{center}
\end{table}

When we use ResNet as the image encoder, some methods seem cannot preserve the improvement. Worth mentioning, SLIP~\cite{mu2021slip} and FILIP~\cite{yao2021filip} do not report the results of ResNet models. We conjecture there are two reasons: 1) ResNet models might need more dedicated hyper parameter tuning. 2). Fine-grained alignment requires the image features to be non-overlapped, which is unachievable for ConvNets. However, DeCLIP can still brings 7.2\% improvement over the CLIP baseline.

\subsection{Model}

While most attention is paid to image encoders, little research is conducted on text encoders. 
Most literature follows the exact setting from CLIP, \ie, a 12-layer transformer. 
Therefore, we expect to study the role of the text encoder, and further explore whether the training efficiency can be improved by reducing the parameters of the text encoder without affecting the performance.

\begin{table}[h]
\begin{center}
\caption{\label{text-encoder-size} Zero-shot top1 accuracy on ImageNet. All models are trained with YFCC15M-V2~\cite{li2021supervision}. The image encoder is ViT-B32, we vary the layer number of transformers in the text encoder.
}
\begin{small}
\begin{tabular}{ccc}
\toprule
\textbf{Method} & \textbf{Layer number} & \textbf{Accuracy} \\  
\midrule
\multirow{4}{*}{CLIP}     & 1                                    & 29.9        \\
                                 & 3                                    & 34.2      \\
                                 & 6                                    & 34.3      \\
                                 & 12                                   & 32.8      \\

\midrule
\multirow{4}{*}{DeFILIP}     & 1                                    & 39.7        \\
                                 & 3                                    & 44.1      \\
                                 & 6                                    &  44.3    \\
                                 & 12                                   & 45.0     \\ 
\bottomrule
\end{tabular}
\end{small}
\end{center}
\vspace{-1.0em} 
\end{table}

As shown in Tab.~\ref{text-encoder-size}, we try 1/3/6/12-layer transformer for CLIP-ViTB32 and DeFILIP-ViTB32. Surprisingly, we find that
(1) For the primitive CLIP method, text encoders with 6 layers of transformers achieve the best results instead of the default 12 layers. A 3-layers transformer is enough to achieve high results. (2) For the DeCLIP, which applies more supervision, the text encoder is more critical. However, even if half the number of layers, it does not significantly affect the final performance. Such an exciting result shows that curtailing the text-encoder is an efficient approach to reducing training costs.

\section{Conclusions} 
In this paper, we propose the first CLIP-benchmark that includes state-of-the-art methods. We benchmark these methods under the same training recipe using the same data. Our CLIP-benchmark also brings some insights about data, supervision, and model. Moreover, we further propose DeFILIP to make a stronger baseline for this task. The CLIP-benchmark would be released to the public for future research. We hope this technical report could avoid duplicate data cleaning efforts and provide a consistent benchmark to facilitate fair comparisons.

{\small
\bibliographystyle{ieee_fullname}
\bibliography{egbib}
}

\end{document}